\documentclass[sigconf]{acmart}
\usepackage{multirow}
\usepackage{balance}
\AtBeginDocument{%
  }

\setcopyright{acmlicensed}
\copyrightyear{2025}
\acmYear{2025}
\setcopyright{acmlicensed}\acmConference[ICMR '25]{Proceedings of the 2025 International Conference on Multimedia Retrieval}{June 30-July 3, 2025}{Chicago, IL, USA}
\acmBooktitle{Proceedings of the 2025 International Conference on Multimedia Retrieval (ICMR '25), June 30-July 3, 2025, Chicago, IL, USA}
\acmDOI{10.1145/3731715.3733403}

\acmISBN{979-8-4007-1877-9/2025/06}

\begin{document}
\title{Multiscale Adaptive Conflict-Balancing Model For Multimedia Deepfake Detection}

\author{Zihan Xiong}
\email{2022090905007@std.uestc.edu.cn}
\orcid{0009-0008-5618-663X}
\affiliation{%
    \institution{University of Electronic Science and Technology of China}
  \city{Chengdu}
  \state{Sichuan}
  \country{China}
}

\author{Xiaohua Wu}
\authornote{Corresponding author}
\email{wuxh@uestc.edu.cn}
\orcid{0009-0007-1220-7219}
\affiliation{%
    \institution{University of Electronic Science and Technology of China}
  \city{Chengdu}
  \state{Sichuan}
  \country{China}
}

\author{Lei Chen}
\email{202322090726@std.uestc.edu.cn}
\orcid{0009-0003-1493-7064}
\affiliation{%
    \institution{University of Electronic Science and Technology of China}
  \city{Chengdu}
  \state{Sichuan}
  \country{China}
}

\author{Fangqi Lou}
\email{202422090730@std.uestc.edu.cn}
\orcid{0009-0007-8925-315X}
\affiliation{%
    \institution{University of Electronic Science and Technology of China}
  \city{Chengdu}
  \state{Sichuan}
  \country{China}
}

\renewcommand{\shortauthors}{Zihan Xiong et al.}

\begin{abstract}
Advances in computer vision and deep learning have blurred the line between deepfakes and authentic media, undermining multimedia credibility through audio-visual forgery. Current multimodal detection methods remain limited by unbalanced learning between modalities. To tackle this issue, we propose an Audio-Visual Joint Learning Method (MACB-DF) to better mitigate modality conflicts and neglect by leveraging contrastive learning to assist in multi-level and cross-modal fusion, thereby fully balancing and exploiting information from each modality. Additionally, we designed an orthogonalization-multimodal pareto module that preserves unimodal information while addressing gradient conflicts in audio-video encoders caused by differing optimization targets of the loss functions. Extensive experiments and ablation studies conducted on mainstream deepfake datasets demonstrate consistent performance gains of our model across key evaluation metrics, achieving an average accuracy of 95.5\% across multiple datasets. Notably, our method exhibits superior cross-dataset generalization capabilities, with absolute improvements of 8.0\% and 7.7\% in ACC scores over the previous best-performing approach when trained on DFDC and tested on DefakeAVMiT and FakeAVCeleb datasets.
\end{abstract}

\begin{CCSXML}
<ccs2012>
<concept>
<concept_id>10010147.10010257</concept_id>
<concept_desc>Computing methodologies~Machine learning</concept_desc>
<concept_significance>500</concept_significance>
</concept>
<concept>
<concept_id>10010147.10010178.10010224</concept_id>
<concept_desc>Computing methodologies~Computer vision</concept_desc>
<concept_significance>500</concept_significance>
</concept>
<concept>
<concept_id>10010147.10010257.10010293</concept_id>
<concept_desc>Computing methodologies~Machine learning approaches</concept_desc>
<concept_significance>500</concept_significance>
</concept>
</ccs2012>
\end{CCSXML}

\ccsdesc[500]{Computing methodologies~Machine learning}
\ccsdesc[500]{Computing methodologies~Computer vision}
\ccsdesc[500]{Computing methodologies~Machine learning approaches}

\keywords{Multimedia machine learning, Video-Audio Deepfake, Multimodal Fusion,  Contrastive Learning}
%%  \includegraphics[width=\textwidth]{sampleteaser}
%%  \caption{Seattle Mariners at Spring Training, 2010.}
%% \Description{Enjoying the baseball game from the third-base
%%  seats. Ichiro Suzuki preparing to bat.}

\maketitle
\section{INTRODDUCTION}
Lately, with the rapid development of computer vision and deep learning, it has become possible to create complex deepfake videos, especially multimodal audio-video deepfakes which have become increasingly rampant\cite{wang2020imaginator, li2020advancing, chan2019everybody}. Audio-video deepfake information spreads rapidly on the internet, not only infringing on personal privacy but also potentially causing social security issues. However, research targeting this form of forgery remains relatively scarce, mostly focusing on unimodal detection without fully utilizing multimodal information\cite{zhou2021face,sabir2019recurrent}, creating significant pain points in areas such as cybersecurity, authentication, and media content review. Therefore, there is an urgent need to develop a detection method that fully balances and utilizes multimodal information to combat these audio-video deepfakes.           

Indeed, research on multimodal deepfake detection is gradually gaining attention\cite{chugh2020not,10233898,nie2024frade}. Existing modal fusion methods, however, remain overly simplistic, often employing linear concatenation that ignores the importance allocation and equally combines features from each modality\cite{mittal2020emotions}. This strategy fails to adequately explore and utilize the potential complementary information between individual modalities and lacks the exploitation of deeper information. During the fusion process, the lack of in-depth feature interaction and cross-modal relationship modeling may lead to bias towards one modality, thereby affecting the overall detection performance. Some modal fusion methods introduce attention mechanisms\cite{cai2022you,zhuang2025vasparseefficientvisualhallucination}, which can dynamically adjust weights based on the importance of different modalities, thus better capturing the relevance and differences among multimodal features and effectively enhancing the comprehensive utilization capability of multimodal information, improving the overall detection performance. The aforementioned methodologies predominantly employ contrastive learning to reduce the distance between similar samples while increasing the separation between dissimilar ones. This approach facilitates a more discriminative representation of features in different modalities, thereby allowing the model to learn comprehensive and balanced characteristics\cite{yin2025atrimitigatingmultilingualaudio,xie24c_interspeech,10243130}. However, even after applying contrastive learning, clustering information can be further exploited to enhance feature representation.

Moreover, incorporating contrastive loss and classification loss for both single-modal and multimodal scenarios introduces complexities during backpropagation, particularly concerning gradient conflicts. If gradients from different modalities conflict during training, it may impede the effective integration of multimodal information, leading to suboptimal performance.
Therefore, it is imperative to investigate a more holistic and efficient method for multimodal fusion that maintains equilibrium throughout the entire training process, we aim to develop a robust multimodal fusion framework that effectively integrates information from various modalities, thereby improving the overall performance and reliability of deepfake detection systems.

Inspired by the above analysis, we propose a multimodal contrastive learning method aimed at reducing biased emphasis during multimodal fusion and addressing non-negligible modality information conflicts when detecting multimodal deepfakes. As shown in Figure~\ref{fig:enter-label}, we first extract modality features from audio and video, and these two modality features undergo clustering through category and authenticity contrastive learning, generating two sets of weights to guide modality fusion. Then, single-modality features are deeply extracted to obtain finer-grained features, which are secondarily fused with the initial fusion information, allowing single modalities to learn global multimodal information. After interacting with shallow and deep fusion information, the final fused feature information is obtained. Finally, we design an orthogonalization-multimodal pareto module to maintain single-modality information while resolving gradient conflicts caused by loss functions with different optimization objectives in audio and video encoders. The main contributions of this paper are summarized as follows:
\begin{itemize}
\item We design a modality-specific and cross-modal contrastive learning module with adaptive temperature regulation based on the authenticity of audio and video (which performs contrastive clustering processing in a unified mapping space) subsequently using video and audio information to generate weight parameters for modality fusion, optimizing the fusion effect of multimodal information.
\item To address gradient conflicts between unimodal and multimodal processes, we propose an orthogonalization-multimo\\dal Pareto optimization module. This approach not only resolves conflicts but also promotes diversity and complementarity in modality fusion.
\item Through experiments, we demonstrate that our designed model achieves optimal performance with accuracies of 96.8\% on DefakeAVMiT, 91.7\% on FakeAVCeleb, and 97.9\% on DFDC, and multiple ablation experiments fully illustrate the effectiveness of the modules we have designed.
\end{itemize}
\section{RELATED WORK}

\subsection{Unimodal Deepfake Detection Methods}
The rapid advancement of synthetic media has significantly compromised cybersecurity frameworks, identity authentication systems, and content moderation protocols. Contemporary detection methodologies are primarily categorized into two paradigms: unimodal approaches that independently analyze individual video sequences for facial inconsistencies or audio tracks for synthetic speech patterns, and multimodal frameworks such as audio-visual cross-modal analysis, which leverage inter-modal correlations to achieve superior detection accuracy.

{\bfseries Image Forgery Detection : }Advancements in deep learning have spurred the development of CNN-based methods. One such approach\cite{li2018fast}introduced a rapid and effective copy-move forgery detection algorithm through hierarchical feature point matching. Another study\cite{mehrjardi2023survey}critically examined the limitations of conventional techniques and explored the efficacy of deep learning methodologies for the detection of forgery. Subsequently, an increasing number of network architectures and modules have emerged \cite{tariq2018detecting,tariq2019gan,gu2022exploiting}, exemplified by the introduction of FCD-Net \cite{han2023fcd}, a novel network framework designed to differentiate various types of deepfake facial images with common origins.

{\bfseries Video Forgery Detection : }The temporal continuity inherent in videos presents both an aid and a challenge for deepfake detection. To improve the performance of the model, advanced techniques can be employed, such as attention mechanisms\cite{Dang_2020_CVPR}or Capsule Networks\cite{nguyen2019capsule} can be employed. Although these methods operate primarily on static frames, their effectiveness can be enhanced by frame-level score aggregation or using different color spaces as input \cite{he2019detection}. Temporal dependencies between frames can be captured using Long Short-Term Memory Networks (LSTM)\cite{sohrawardi2019poster} , thereby improving detection accuracy. In pursuit of enhanced generalization capabilities, hierarchical neural networks were leveraged in \cite{fernando2019exploiting} to exploit long-term dependencies, integrating attention mechanisms and adversarial training to fortify robustness against compressed videos and novel forgeries.

\begin{figure*}[t] 
    \centering
    \includegraphics[width=\textwidth]{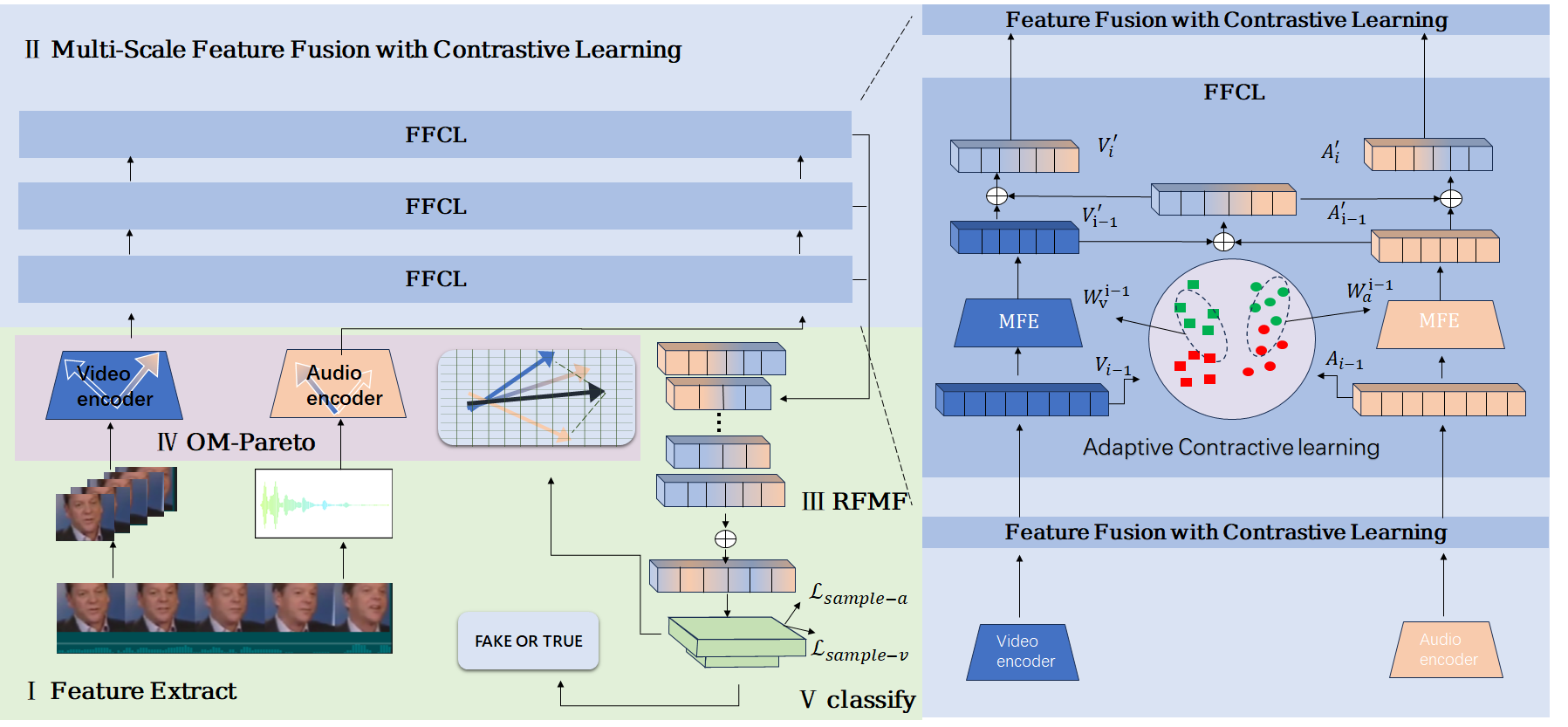}
    \caption{Multiscale Adaptive Conflict-Balancing model for multimedia Deepfake detection deptly balances conflicts and biases among modalities, thereby fully unlocking the potential of intra-modality and inter-modality information. There are five stages in this method: Feature Extract, Multi-Scale Feature Fusion With Contrastive Learning, RFMF, OM-Pareto, Classify}
    \label{fig:enter-label}
\end{figure*}

%\subsubsection{Audio Forgery Detection}
{\bfseries Audio Forgery Detection : } Deepfake audio methodologies predominantly focus on voice conversion \cite{malik2020derivative} and Text-to-Speech (TTS), also known as speech synthesis. Some audio deepfakes involve selectively altering specific segments of original clips using AI technologies while preserving the overall authenticity of the recording, rather than synthesizing entire audio tracks. Current approaches to detecting deepfake audio facts largely focus on analyzing acoustic properties, such as speech patterns, vocal characteristics, and linguistic nuances, to identify anomalies that deviate from natural audio attributes. Furthermore, leveraging aligned video information, such as lip movement synchronization with speech\cite{huang2022fastdiff}, inconsistencies between background environments in videos and spoken content, or discrepancies between audio narratives and video scenarios, can further aid in identifying forged audio.

\subsection{Unimodal Bias in Multimodal Deep Learning}

The success of multimodal deep learning hinges on the effective integration and utilization of multiple modalities \cite{8269806,liang2024foundations,ru2025reallyfilterrandomnoise}. However, during joint training, certain multimodal networks exhibit a propensity to overly rely on modalities that are easier or faster to learn, thereby neglecting other modalities \cite{lin2017focal,cadene2019rubi,wang2020makes}.\cite{wu2022characterizing} introduced a gradient analysis-based metric to evaluate unimodal bias within multimodal models. This approach quantifies how much each modality contributes to the final prediction, revealing any potential biases towards specific modalities.\cite{peng2022balanced}empirically demonstrated that competition among modalities can lead to the neglect of certain modalities. Their findings underscore the importance of balancing contributions from different modalities to ensure comprehensive learning\cite{10.1007/978-3-031-72754-2_18,zhuang2025vargptv11improvevisualautoregressive}.

To address these challenges, our work introduces a learnable contrastive learning framework aimed at facilitating more balanced and holistic feature representation. This method not only aids the model in acquiring a richer and more equitable understanding of the data but also guides modality fusion effectively. Additionally, we employ an Orthogonalization-multimodal Pareto formulation to mitigate conflicts arising from gradient updates across multiple modalities, ensuring that no single modality dominates the learning process.

\section{METHOD}
We propose a novel Multiscale Adaptive Conflict-Balancing model for multimedia Deepfake detection (MACB-DF) in Figure~\ref{fig:enter-label}. This model adeptly balances conflicts and biases among modalities, thereby fully unlocking the potential of intra-modality and inter-modality information.

\subsection{Feature Extract}
The MACB-DF model processes video-audio pairs $(V, A) \sim P$ through two encoders: a video encoder $E_v$ and an audio encoder $E_a$. For video input $V$, temporal and spatial information is encoded by $E_v$ via self-attention layers and feed-forward networks, producing a global spatiotemporal video feature $H_v^{(L)}$. 
Audio processing begins with the STFT of signal $A$, generating a spectrogram $S(t, f)$ where $t$ indexes time frames and $f$ denotes frequency. This spectrogram is then projected to Mel-scale using filter banks $\{H_m(f)\}_{m=1}^F$:
\begin{equation}
M(t, m) = \sum_f H_m(f) |S(t, f)|^2
\end{equation}
where $F$ is the number of Mel filters. The logarithmic transformation $\log M(t, m)$ is applied to amplify low-energy components. The resulting log Mel-spectrum is encoded by $E_a$ through self-attention mechanisms and feed-forward operations, yielding a global audio feature $H_a^{(L)}$ with temporal dependencies.

\subsection{Multi-Scale Feature Fusion With Contrastive Learning}

\subsubsection{Multi-modal Adaptive Contrast Learning}

To facilitate more effective exploitation of the correlations between video and audio by the two single-modal encoders, we employ contrastive learning to align video and audio embeddings. Traditional inter-modality contrastive learning methods, however, often suffer from an inability to fully leverage the intrinsic information within each modality. Moreover, fluctuations in the temperature parameter can significantly impact the gradient of the contrastive loss, leading either to insufficient capture or excessive blurring of detailed information. Additionally, these methods do not make full use of the information post-clustering.

\begin{figure}[h]
  \centering
    \includegraphics[width=0.7\linewidth]{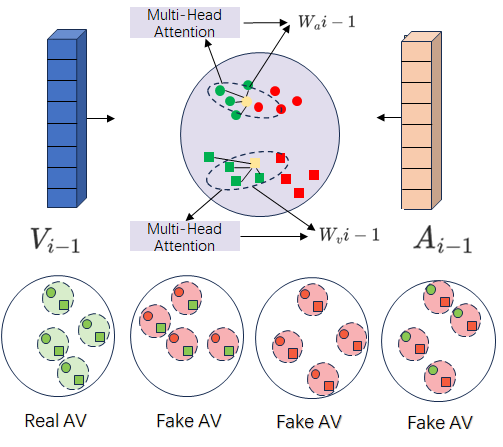}
      \caption{Adaptive Contrast Learning}
        \label{Adaptive Contrast Learning}
\end{figure}
        
Therefore, we propose \textit{Multi-modal Adaptive Contrast Learning} (MACL),
To formalize the contrastive loss from audio to video, we define it as follows:
\begin{equation}
\begin{split}
L_{\text{av}}& = - \mathbb{E}_{p(A,V)}\\&
\log \frac{\exp\left( \text{logits}_{\text{pos}_i} \right)}{\exp\left( \text{logits}_{\text{pos}_i} \right) + \sum_{k=1}^{K} \exp\left( \text{logits}_{\text{neg}_{ik}} \right)}
\end{split}
\end{equation}
The logits for positive pairs are computed as:$\frac{S(x_{a_i}, x_{v_i})}{\tau}$while those for negative pairs incorporate a margin $m$:$ \frac{S(x_{a_i}, x_{v_k}^{-})}{\tau} - \frac{m}{\tau}.$

For each pair of samples within a batch, we define their attention weights as:
\begin{equation}
w_{ij} = \frac{\exp(f_\theta(s_{ij}))}{\sum_{k,l} \exp(f_\theta(s_{kl}))}
\end{equation}
Where $f_\theta$ represents a parameterized attention network that considers features such as the similarity value itself, deviation from the batch mean, and the distance between sample pairs in the feature space.

By integrating Bayesian estimation with the attention mechanism, we derive the updated formulation for the temperature parameter:
\begin{equation}
\tau = \tau_0 \cdot \exp\left(\sum_{k=1}^K \beta_k \phi_k(S)\right)
\end{equation}
where $\phi_k(S)$ are a set of basis functions including: $\phi_1(S) = \text{Var}(S)$ (the variance term), $\phi_2(S) = \sum_{i,j} w_{ij}|s_{ij} - \mathbb{E}[S]|^3$ (weighted skewness), and $\phi_3(S) = H(w)$ (entropy of attention weights). The coefficients $\beta_k$ are learnable parameters that combine these features.

A temporal adjustment factor is introduced to adapt the temperature parameter over time:
\begin{equation}
\gamma_t = \sigma(\text{MLP}([h_{t-1}, \Delta\tau_{t-1}, \text{grad}_{\tau} L_{t-1}]))
\end{equation}
Consequently, the final temperature parameter at time $t$ is defined as:
\begin{equation}
\tau_t = \gamma_t \cdot \tau_{t-1} + (1-\gamma_t) \cdot \tau_{\text{new}}
\end{equation}
where $h_{t-1}$ encodes historical states, $\Delta\tau_{t-1}$ represents the temperature change from the previous step, and $\text{grad}_{\tau} L_{t-1}$ indicates the gradient of the loss with respect to the temperature. This approach leverages Bayesian inference for theoretical guarantees, captures fine-grained characteristics through the attention mechanism, and achieves temporal adaptation via dynamic adjustments.

Similarly, the contrastive loss from video to audio is formulated as:
\begin{equation}
\begin{split}
L_{\text{va}} =& - \mathbb{E}_{p(V,A)} \\&
\sum_{i=1}^{N} 
\log \frac{
  \exp\left( \frac{\text{sim}(x_v^i, x_a^i)}{\tau} \right)
}{
  \exp\left( \frac{\text{sim}(x_v^i, x_a^i)}{\tau} \right) + 
  \sum\limits_{k=1}^{K} 
  \exp\left( \frac{\text{sim}(x_v^i, x_{a_k}^{-})}{\tau} - m \right)
}
\end{split}
\label{eq:lva}
\end{equation}
where $x_{a_k}^{-}$ represents the queue of the nearest K negative audio samples not matching with video $V$.

Inspired by this, to preserve the efficacy of information within each single modality, we further implement intra-modality contrastive learning within both video and audio. We aggregate all losses into an adaptive modality contrastive learning loss as follows:
\begin{equation}
L_{C} = \frac{1}{4} \left[ L_{va} + L_{av} + L_{vv} + L_{aa}\right],
\label{eq:lmac}
\end{equation}
where $L_{vv}$ and $L_{aa}$ denote the intra-modality contrastive losses for video and audio, respectively.

\subsubsection{Adaptive Parameter Optimization for Contrastive Learning}

Going further, as shown in Figure~\ref{Adaptive Contrast Learning}, we devise a weight parameter that guides the fusion of two modalities by exploiting the clustering information obtained after contrastive learning. This approach maximizes the utilization of latent information within each modality while optimizing the fusion strategy to enhance complementarity between modalities. As a result, the fused features more effectively represent the authenticity of samples.

We employ an adaptive clustering method that dynamically adjusts the number of clusters $K$ based on data complexity. For video features $\{ x_{v_i} \}$, we obtain cluster centers $\{ c_v^k \}_{k=1}^{K_v}$; for audio features $\{ x_{a_i} \}$, we obtain cluster centers $\{ c_a^k \}_{k=1}^{K_a}$. Incorporating both Mahalanobis distance and cosine similarity, we propose a composite distance measure:
\begin{align}
D_{v_i}= \beta \cdot \frac{d_{k_i}}{d_k^{\text{max}}} + (1 - \beta) \cdot s_{k_i}
\end{align}

where:
$ d_{k_i} = \sqrt{(x_{k_i} - c_{k_i})^\top \Sigma_k^{-1} (x_{k_i} - c_{k_i})} $, 
$ s_{k_i} = 1 - \frac{x_{k_i}^\top c_{k_i}}{\| x_{k_i} \| \| c_{k_i} \|} $ ,$k \in \{a, v\}$, effectively measure the true distance between a sample and the cluster center, $\beta$ is a balancing parameter, and $d_v^{\text{max}}$, $d_a^{\text{max}}$ are the maximum Mahalanobis distances in their respective modalities, used for normalization. The composite distance $D_{v_i}$ captures both statistical distribution differences and directional discrepancies between samples and cluster centers. By introducing the balancing parameter $\beta$, we adjust the contributions of these two distances, thus obtaining a comprehensive evaluation of the relationship between samples and cluster centers. This enhanced distance metric improves clustering accuracy and provides a more reliable basis for modality fusion, thereby strengthening the representation of sample authenticity in the fused features.

Let $m \in \{v, a\}$ where $v$ denotes the video modality and $a$ the audio modality. We then apply a multi-head attention mechanism to comprehensively capture feature associations. Finally, we derive the importance scores:
\begin{equation}
w_{m_i} = \gamma \cdot \text{ReLU}\left( f_m( \text{MHAttention}_m ) \right) + (1 - \gamma) \cdot \frac{1}{D_{m_i} + \epsilon}
\end{equation}

\begin{figure}[h]
  \centering
    \includegraphics[width=\linewidth]{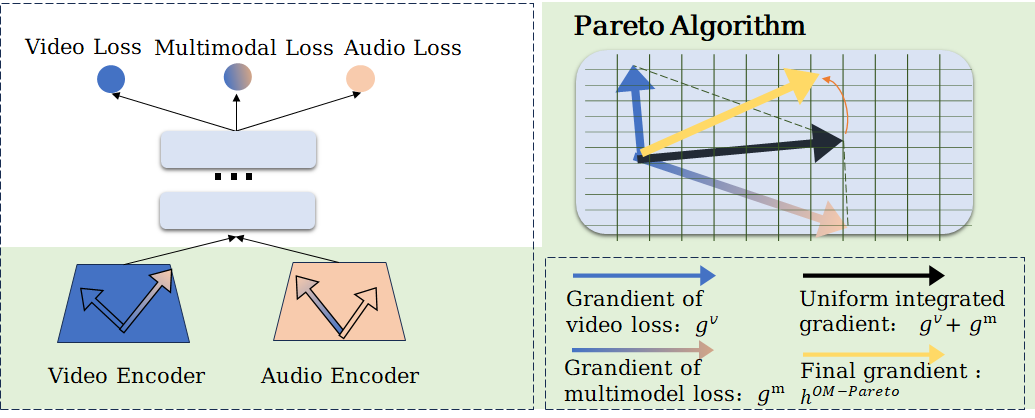}
      \caption{Orthogonalization-multimodal Pareto}
        \label{Orthogonalization-multimodal Pareto}
\end{figure}

where $f_m$ is a learnable non-linear mapping function for modality $m$, $\gamma \in [0,1]$ is a balance parameter controlling the weights of the attention output and distance measures.

\subsubsection{Multi-Scale Feature Extraction}

After encoding, audio and video yield feature representations $V_i$ and $A_i$. Inspired by large-kernel convolutions, we design modality-specific feature extractors for video ($V_{i-1}$) and audio ($A_{i-1}$), capturing long-range dependencies and multi-scale information.

For video, the Multi-Scale Spatio-Temporal Large Kernel Attention (MSTLKA) block processes input features $H_v^{(L)} \in \mathbb{R}^{T \times C \times H \times W}$ as follows:
\begin{align}
N &= \text{LN}(H_v^{(L)}) \\ 
H_v^{(L)} &= H_v^{(L)} + \lambda_1 f_3(\text{MSTLKA}(f_1(N)) \otimes f_2(N)) \\ 
N &= \text{LN}(H_v^{(L)}) \\ 
H_v^{(L)} &= H_v^{(L)} + \lambda_2 f_6(\text{GSAU-video}(f_4(N), f_5(N)))
\end{align}
Here, $f_i(\cdot)$ are $1 \times 1$ convolutions, and $\lambda_1$, $\lambda_2$ are learnable scalars.GSAU-video uses 3D depthwise convolutions for gating.

\textbf{MSTLKA} extends LKA to 3D convolutions for spatiotemporal dependencies. STLKA is:
\begin{equation}
\text{STLKA}(X) = f_{PW}(f_{DWD}(f_{DW}(X)))
\end{equation}
where:
$f_{DW}$: 3D depthwise convolution $(2d-1) \times (2d-1) \times (2d-1)$.
$f_{DWD}$: 3D dilated depthwise convolution $\lceil \frac{K}{d} \rceil$.
$f_{PW}$: Pointwise convolution.

Input features are split into $n$ groups, and MSTLKA is applied at different scales:
\begin{equation}
\text{MSTLKA}_i(X_i) = G_i(X_i) \otimes \text{STLKA}_i(X_i)
\end{equation}
$G_i(\cdot)$ is a 3D depthwise convolution gating mechanism, dynamically adjusting attention.

The audio pathway shares the same structure as video but employs 2D operations for time-frequency modeling, with MSTLKA replaced by MFTLKA and GSAU simplified to 2D.

\subsubsection{Multimodal Feature Fusion}

Utilizing the computed fusion weights, we perform a weighted fusion of deep video and audio features to obtain a fused feature representation. Specifically, the calculation for the fused feature $x_{\text{fused}_i}$ is defined as:
\begin{equation}
x_{\text{fused}_i} = w_{v_i} \cdot V_{i-1} + w_{a_i} \cdot A_{i-1}
\end{equation}
In this equation, $w_{v_i}$ and $w_{a_i}$ denote the fusion weights for the video and audio modalities, respectively, while $V_{i-1}$ and $A_{i-1}$ represent the deep features from the video and audio streams at the previous layer.

By employing this method, we effectively integrate information from both video and audio modalities, leveraging their complementary nature to enhance the model's capability in representing multimodal data. Subsequently, using $x_{\text{fused}_i}$—which encapsulates global audio-video fusion information—we guide the learning of complementary modality information and global context in deeper layers:
\begin{equation}
K_{i}= K'_{i-1} \otimes x_{\text{fused}_i} \quad K \in \{V, A\}
\end{equation}

\subsection{Representation Fusion via Multi-Head Attention Framework}

After multi-scale feature fusion, the audio and video modalities are represented as $\{ha_i, hv_i... ha_{i+n}, hv_{i+n}\}$, which are fused into a joint vector for deepfake detection. Our fusion mechanism employs a Transformer-based self-attention framework:
$\text{head}_i = \text{Attention}(Q\\W^q_i, KW^k_i, VW^v_i)$Stack the representations into matrix $M = [ha_i, hv_i\\...ha_{i+n}, hv_{i+n}]$ and process through multi-head attention:  
$M = (\text{head}_1 \oplus \cdots \oplus \text{head}_n)W^o$
where $\Theta_{\text{att}}$ denotes attention parameters and $W^o$ is the output transformation matrix. This enables cross-modal interaction through multiple attention heads, producing enriched features $M$ for detection.

\subsection{Classify}

The classification module operates at both frame-level and sample-level granularity.

Utilizing the fused feature matrix \( M \) output by the RFMF module, an MLP is designed for authenticity discrimination. The model adopts a progressive dimensionality reduction architecture: stacked fully connected layers gradually compress feature dimensions, stepwise abstracting discriminative high-level semantic representations. Between network layers, a hybrid integration of BatchNorm and Dropout layers implements dual regularization — standardizing feature distributions through BatchNorm while randomly deactivating neurons via Dropout — thereby mitigating the risk of overfitting. The final layer employs the Sigmoid function to map features into the \([0,1]\) interval, outputting the estimated probability of a sample being forged. The training process utilizes the binary cross-entropy loss function:
\begin{equation}
\mathcal{L}_{\text{sample}}^k = -\frac{1}{N}\sum_{i=1}^N \left[ y_i \log \hat{y}_i + (1 - y_i) \log(1 - \hat{y}_i) \right]
\end{equation}

where \( k \in \{audio, video\} \), \( y_i \in \{0, 1\} \) denotes the ground-truth label and \( \hat{y}_i \) represents the predicted probability. 

For each temporal frame $t$ in feature after Multi-Scale Feature Fusion With Contrastive Learning, independent MLP branches process audio/video features:
\begin{align}
\mathcal{L}_{\text{frame}}^k &= -\frac{1}{N}\sum_{i=1}^N \left[ y_i \log \hat{y}_i + (1 - y_i) \log(1 - \hat{y}_i) \right]
\end{align}
where $k \in \{\text{audio}, \text{video}\}$.

\begin{table*}[htbp]
\centering
\caption{COMPARISON WITH OTHER METHODS ON INTRA-DATASETS USING SAMPLE-LEVEL AUC AND ACC METRICS }
\label{tab:COMPARISON WITH OTHER METHODS ON INTRA-DATASETS USING SAMPLE-LEVEL AUC AND ACC METRICS }
\begin{tabular}{@{}llcccccccc@{}}
\toprule
\multirow{2}{*}{Methods} & \multirow{2}{*}{Modality} & \multicolumn{2}{c}{DefakeAVMiT} & \multicolumn{2}{c}{FakeAVCeleb} & \multicolumn{2}{c}{DFDC} & \multicolumn{2}{c}{Avg}\\
\cmidrule(lr){3-4} \cmidrule(lr){5-6} \cmidrule(l){7-8} \cmidrule(l){9-10}
& & ACC(\%) & AUC(\%) & ACC(\%) & AUC(\%) & ACC(\%) & AUC(\%) & ACC(\%) & AUC(\%)\\
\midrule
Multiple-Attention \cite{zhao2021multiattentionaldeepfakedetection} (2021) & visual & 85.1 & 87.3 & 77.6 & 79.3 & 82.5 & 84.8 & 81.7 & 83.8 \\
SLADD \cite{9879418} (2022) & visual & 73.3 & 77.5 & 70.5 & 72.1 & 73.6 & 75.2 & 72.5 & 74.9 \\
\midrule
ECAPA-TDNN \cite{Desplanques_2020} (2020) & audio & 70.2 & 72.6 & 59.8 & 62.7 & 67.3 & 69.8 & 65.8 & 68.4 \\
AASIST \cite{jung2022aasist} (2021) & audio & 67.5 & 69.8 & 53.4 & 55.1 & 64.8 & 68.4 & 61.9 & 64.4 \\
\midrule
AVFakeNet \cite{ilyas2023avfakenet} (2022) & audio-visual & 91.8 & 93.7 & 78.4 & 83.4 & 82.8 & 86.2 & 84.3 & 87.8 \\
VFD \cite{cheng2023voice} (2022) & audio-visual & 93.4 & 95.6 & 81.5 & 86.1 & 80.9 & 85.1 & 85.3 & 88.9 \\
BA-TFD \cite{cai2022you} (2022) & audio-visual & 92.1 & 94.9 & 80.8 & 84.9 & 79.1 & 84.6 & 84.0 & 88.1 \\
AVoiD-DF \cite{10081373} (2023) & audio-visual & 95.3 & 97.6 & 83.7 & 89.2 & 91.4 & 94.8 & 90.1 & 93.9 \\
MCL \cite{10243082} (2024) & audio-visual & - & - & 85.9 & 89.2 & 97.5 & 98.3 & 91.7 & 93.8 \\
\midrule
\textbf{MACB-DF (Ours)} & \textbf{audio-visual} & \textbf{96.8} & \textbf{98.7} & \textbf{91.7} & \textbf{93.2} & \textbf{97.9} & \textbf{98.8} & \textbf{95.5} & \textbf{96.9} \\
\bottomrule
\end{tabular}
\end{table*}

\subsection{Orthogonalization-multimodal Pareto}
consider the following constrained optimization problem:
\begin{equation}
\begin{aligned}
\min_{\alpha_m,\alpha_u} \quad & \mathcal{L}_0 = \left\| \alpha_m \mathbf{g}_m + \alpha_u \mathbf{g}_u \right\|^2 \\
\text{s.t.} \quad & \alpha_m + \alpha_u = 1 \\
& \alpha_m, \alpha_u \geq 0
\end{aligned}
\label{eq:original_problem}
\end{equation}
where $\mathbf{g}_m \in \mathbb{R}^d$ and $\mathbf{g}_u \in \mathbb{R}^d$ represent multimodal and unimodal gradient vectors, respectively.

An implicit orthogonal regularization term is introduced during gradient updates to construct augmented gradients:
\begin{equation}
\nabla_{\alpha}\mathcal{L}_{\text{aug}} = \nabla_{\alpha}\mathcal{L}_0 + \lambda_{\text{orth}} \cdot \nabla_{\alpha} \left( \alpha_m \alpha_u |\mathbf{g}_m^\top \mathbf{g}_u| \right)
\label{eq:augmented_grad}
\end{equation}

The regularization strength parameter is adaptively adjusted based on gradient directional similarity:
\begin{equation}
\lambda_{\text{orth}} = \frac{\lambda_0}{1 + \exp(\kappa \cos\theta)}, \quad \cos\theta = \frac{\mathbf{g}_m^\top \mathbf{g}_u}{\|\mathbf{g}_m\| \|\mathbf{g}_u\|}
\label{eq:adaptive_lambda}
\end{equation}
where $\kappa > 0$ denotes the curvature control parameter and $\lambda_0$ represents the baseline regularization strength.

Non-Conflict Scenario ($\cos\theta \geq 0$)
\begin{itemize}
\item Maintain original Pareto uniform weighting: $\alpha_m = \alpha_u = 0.5$
\item Apply mild orthogonal correction: $\lambda_{\text{orth}} \leftarrow 0.1\lambda_0$
\end{itemize}

Conflict Scenario ($\cos\theta < 0$)
\begin{itemize}
\item Execute complete PGD iterations
\item Noise-amplified gradient magnitude:
\begin{equation}
\mathbf{h}_{\text{final}} = \frac{\alpha_m^* \mathbf{g}_m + \alpha_u^* \mathbf{g}_u}{\|\alpha_m^* \mathbf{g}_m + \alpha_u^* \mathbf{g}_u\|} \cdot \left(1 + \frac{|\cos\theta|}{1 + \|\mathbf{g}_m\|/\|\mathbf{g}_u\|}\right) \|\mathbf{g}_m + \mathbf{g}_u\|
\label{eq:noise_boost}
\end{equation}
\end{itemize}

\section{EXPERIMENTS}
\subsection{Datasets}

\textbf{DefakeAVMiT}
This multimodal dataset \cite{10081373} contains 6,480 audiovisual samples generated using 8 distinct deepfake methods. It features four combinatorial classes:
\textbf{V\textsubscript{R}A\textsubscript{R}}: Real visual/audio ,
\textbf{V\textsubscript{F}A\textsubscript{R}},\textbf{V\textsubscript{R}A\textsubscript{F}}, \textbf{V\textsubscript{F}A\textsubscript{F}}: Hybrid/fake combinations .Samples are temporally aligned to ensure audiovisual synchronization.

\begin{table}
  \caption{ABLATION STUDY ON CONTRASTIVE LEARNING }
  \label{tab:ABLATION STUDY ON CONTRASTIVE LEARNING }
  \begin{tabular}{ccccl}
    \toprule
\multirow{2}{*}{Methods} & \multicolumn{2}{c}{FakeAVCeleb} & \multicolumn{2}{c}{DFDC} \\
\cmidrule(lr){2-3} \cmidrule(lr){4-5}
&  ACC(\%) & AUC(\%) & ACC(\%) & AUC(\%)  \\
    \midrule
    MACB-DF w/o $L_{C}$ & 88.5& 90.2 & 93.2 & 95.1\\
    MACB-DF w/o $L_{intra}$ & 90.1& 92.2 & 95.5 & 96.8\\
    MACB-DF w/o $L_{cross}$ & 90.9& 92.8 & 96.9 & 97.2\\
    MACB-DF w/o $w$ & 91.2& 92.8 & 97.3 & 98.4\\
    \midrule
    \textbf{MACB-DF}  & \textbf{91.7} & \textbf{93.2} & \textbf{97.9} & \textbf{98.8} \\  
  \bottomrule
\end{tabular}
\end{table}

\textbf{FakeAVCeleb}
Derived from VoxCeleb2, this benchmark \cite{DBLP:journals/corr/abs-2108-05080} expands 490 real videos to 19,500 synthetic samples through: Visual manipulation: Faceswap/FSGAN, Audiovisual synthesis: SV2TTS/W\\av2Lip.Emphasizes lip-synced audio-visual coherence through frame-level alignment.

\textbf{DFDC}
This large-scale dataset \cite{dolhansky2020deepfake} contains >100k clips from 3,426 subjects featuring: Multi-method visual forgery (DFAE, StyleGAN, etc.),Single-mode audio synthesis (TTS Skins). Notably lacks lip-sync between synthetic audio and visual frames.

\textbf{LAV-DF} 
This large-scale dataset \cite{cai2022you} contains 136,304 videos featuring 153 unique identities, with 36,431 completely real videos and 99,873 videos containing fake segments. The dataset is tasks with per-frame authenticity labels. 

\subsection{Performance}

\subsubsection{sample-level detection performance}

This section evaluates our framework's sample-level detection performance across all training datasets. For rigorous validation, we compare Deepfake detection methods: video-based approaches \cite{zhao2021multiattentionaldeepfakedetection,9879418}, audio-centric methods \cite{Desplanques_2020,jung2022aasist}, and multimodal fusion techniques \cite{ilyas2023avfakenet,cheng2023voice,cai2022you,10081373,10243082}. Datasets were partitioned using randomized stratification (80:20 train-test ratio) to ensure unbiased evaluation.

\begin{figure}[h]
  \centering
  \includegraphics[width=\linewidth]{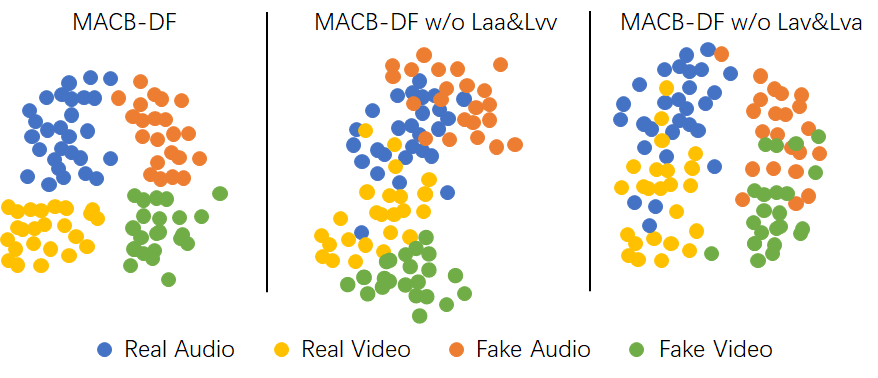}
  \caption{Ablation Study on Contrastive Learning: Latent Space Visualization}
  \label{Ablation Study on Contrastive Learning: Latent Space Visualization}
\end{figure}

\begin{figure}[h]
  \centering
  \includegraphics[width=0.9\linewidth]{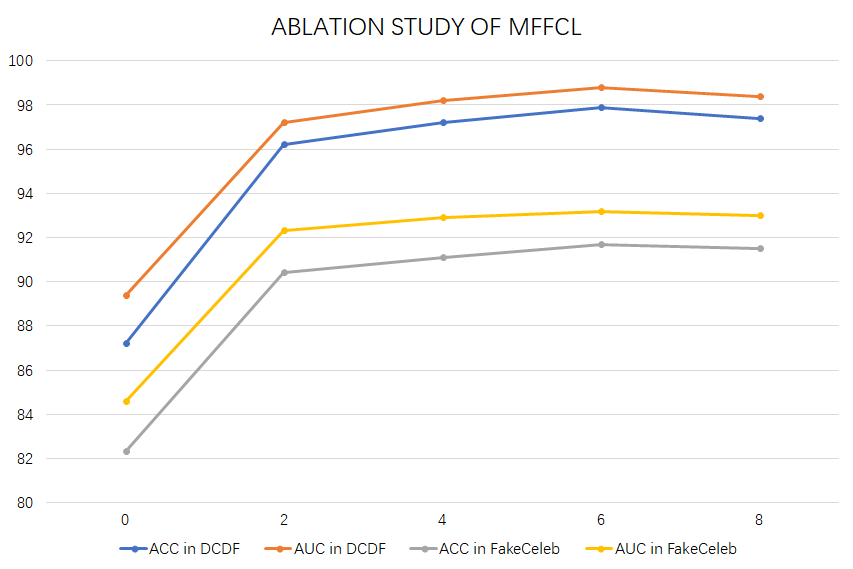}
  \caption{Ablation Study on MFFCL: Impact on AUC/ACC Metrics for DFDC and FakeAVCeleb Datasets}
  \label{Ablation Study on MFFCL: Impact on AUC/ACC Metrics for DFDC and FakeAVCeleb Datasets}
\end{figure}
\begin{figure}[h]
  \centering
  \includegraphics[width=0.9\linewidth]{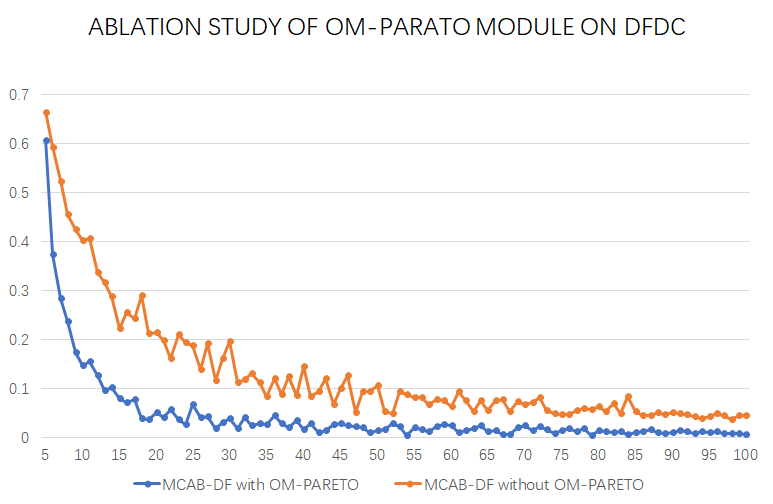}
  \caption{Ablation Study on OM-Pareto: Impact of Loss Function}
  \label{Ablation Study on OM-Pareto: Impact of Loss Function}
\end{figure}
As shown in Table~\ref{tab:COMPARISON WITH OTHER METHODS ON INTRA-DATASETS USING SAMPLE-LEVEL AUC AND ACC METRICS }, MACB-DF demonstrates consistently strong performance across all evaluated datasets. On DefakeAVMiT, it attains 96.8\% ACC and 98.7\% AUC, surpassing existing methods by a significant margin. For FakeAVCele, which features diverse Deepfake variants, MACB-DF demonstrates exceptional performance with 91.7\% ACC and 93.2\% AUC, achieving substantial improvements over prior approaches. On the challenging DFDC dataset, MACB-DF achieves 97.9\% ACC and 98.8\% AUC.  Across all datasets, MACB-DF achieves an average of 95.5\% ACC and 96.9\% AUC, significantly outperforming other state-of-the-art methods. 
 \begin{figure}[h]
  \centering
  \includegraphics[width=0.9\linewidth]{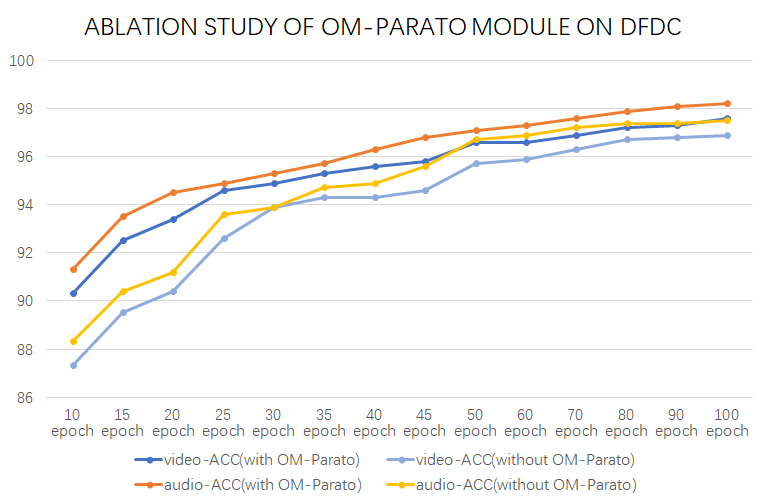}
  \caption{Ablation Study of OM-Pareto: Impact on Video and Audio AUC/ACC Metrics}
  \label{Ablation Study of OM-Pareto: Impact on Video and Audio AUC/ACC Metrics}
\end{figure}
Three key innovations drive this advancement: 1) \textbf{Contrastive Multimodal Alignment}: Reduces biased dominance and information conflicts during fusion 
through contrastive learning; 2) \textbf{Hierarchical Feature Extraction}: Combines global context with deep local patterns through multi-scale architecture.

\section{ABLATION STUDY}
\subsection{Contrastive Learning}
To validate the contribution of contrastive learning, we analyze four variants of our MACB-DF framework with results in Table~\ref{tab:ABLATION STUDY ON CONTRASTIVE LEARNING }: 1) MACB-DF w/o $L_{C}$: Exclusion of contrastive learning mechanisms (ACC drops by 3.2\%/4.7\% on FakeAVCeleb/DFDC compared to full model), 2) MACB-DF w/o $L_{\text{intra}}$: Ablation of contrastive loss $L_{aa}$ and $L_{vv}$ (AUC decreases 1.0\%/2.0\% demonstrating modality-specific alignment necessity), 3)  MACB-DF w/o $L_{\text{cross}}$: Ablation of contrastive loss $L_{av}$ and $L_{va}$ (ACC gap of 0.8\%/1.0\% highlights cross-modal synchronization value), 4) MACB-DF w/o $w$: Removal of adaptive fusion weighting (97.3\% $\rightarrow$ 97.9\% ACC on DFDC proves dynamic weighting superiority). The latent
space visualization of ablation study on contrastive learning is shown in Figure~\ref{Ablation Study on Contrastive Learning: Latent Space Visualization}.
The empirical findings substantiate that contrastive learning significantly enhances discriminative capacity through three synergistic effects: 1) \textbf{Inter-modal Discrepancy Learning}: Intra- and cross-modal contrastive learning enables comprehensive exploitation of inter-modal discrepancies, particularly effective for detecting subtle audio-visual asynchrony, 2) \textbf{Representation Polarization}: The push-pull dynamics between positive pairs (intra-class) and negative pairs (inter-class) amplify decision boundaries by a\% compared to baseline models, 3) \textbf{Cluster-aware Fusion}: Cluster density-based weighting automatically balances modality-specific contributions.

\subsection{Multi-Scale Feature Fusion With Contrastive Learning}
We conducted experiments with five different depths of deep feature extraction layers (0, 2, 4, 6, 8) on the DFDC dataset and FakeAVCeleb dataset. The results are shown in Figure~\ref{Ablation Study on MFFCL: Impact on AUC/ACC Metrics for DFDC and FakeAVCeleb Datasets}. As the number of layers increases, the accuracy first improves and then slightly declines. Notably, the accuracy peaks at six layers, achieving an ACC of 97.9\% on the DFDC dataset and 91.7\% on the FakeAVCeleb dataset. A potential explanation for this phenomenon is that an appropriate number of layers allows the model to effectively fuse deep-level features with global features, thereby enhancing the learning of multimodal information. However, excessive layers may lead to the loss of critical information during the extraction process, resulting in reduced accuracy.

\subsection{Orthogonalization-multimodal Pareto}
For the orthogonalization-multimodal pareto module, its primary function lies in mitigating multi-modal gradient conflicts. To validate this, we evaluate the model accuracy with and without the module across different training epochs, as illustrated in Figure~\ref{Ablation Study on OM-Pareto: Impact of Loss Function} and Figure~\ref{Ablation Study of OM-Pareto: Impact on Video and Audio AUC/ACC Metrics}. The experimental results demonstrate that during the early training stages with fewer epochs, the model equipped with the orthogonalization-multimodal pareto module exhibits a significantly faster loss reduction rate and a more stable convergence trajectory. Moreover, the accuracy of the orthogonalization-multimodal pareto model substantially outperforms its counterpart without the module across both video and audio modalities. Notably, the performance hole gradually diminishes as the number of training epochs increases. These observations conclusively substantiate that our proposed Pareto module effectively alleviates gradient conflicts, thereby facilitating more efficient learning of multi-modal representations during the critical initial training phases.

\begin{table}
  \caption{THE ACCS OF CROSS-DATASET EXPERIMENTS. THE TRAINING SETS AND TESTING SETS FOR CROSS-DATASET}
  \label{tab:THE ACCS OF CROSS-DATASET EXPERIMENTS. THE TRAINING SETS AND TESTING SETS FOR CROSS-DATASET}
  \begin{tabular}{ccl}
    \toprule
\multirow{2}{*}{Methods} & \multicolumn{2}{c}{DFDC}  \\
\cmidrule(lr){2-3} 
& DefakeAVMiT & FakeAVCeleb   \\
    \midrule
    CViT & 47.2& 45.5 \\
    MesoNet & 57.7 & 54.1\\
    MDS & 76.5 & 72.9\\
    AVoiD-DF & 84.4 & 82.8\\
    \midrule
    \textbf{MACB-DF}  & \textbf{91.2} & \textbf{89.2} \\  
  \bottomrule
\end{tabular}
\end{table}

\section{GENERALIZATION TESTS}
The cross-dataset generalization capability of our method is comprehensively evaluated through rigorous experiments, as shown in Table~\ref{tab:THE ACCS OF CROSS-DATASET EXPERIMENTS. THE TRAINING SETS AND TESTING SETS FOR CROSS-DATASET}. When trained on DFDC and tested on DefakeAVMiT and FakeAVCeleb datasets, MACB-DF achieves remarkable ACC scores of 91.2\% and 89.2\% respectively, significantly outperforming existing state-of-the-art methods. Notably, our approach demonstrates 6.8\% and 6.4\% absolute improvements over the previous best method (AVoiD-DF) on two testing sets, revealing superior adaptability to unknown data distributions. This enhanced generalization stems from our balanced multimodal fusion strategy and effective mitigation of modality bias, which prevents overfitting to dataset-specific artifacts. 

\section{CONCLUSION}
We propose MACB-DF, a novel audio-visual joint learning framework for multimedia deepfake detection that addresses modality bias and gradient conflicts in multimodal fusion. By integrating hierarchical contrastive learning with authenticity-aware adaptation, MACB-DF achieves balanced cross-modal representation while preserving fine-grained unimodal features. The Pareto module further resolves optimization conflicts, enhancing robustness and complementarity. Extensive experiments demonstrate state-of-the-art performance on benchmark datasets, with significant improvements in cross-dataset generalization. These results underscore MACB-DF’s effectiveness in critical multimedia applications such as cybersecurity, content authentication, and media review systems. Future work will explore lightweight deployment strategies and temporal consistency modeling to enhance scalability and real-time applicability in dynamic multimedia environments.

\section{ACKNOWLEDGMENTS}
This work was supported by the National Natural Science Foundation of China(62471090), Sichuan Provincial Natural Science Foundation Project(23NSFSC0422), Central University Fund(ZYGX2024Z\\016).

\bibliographystyle{ACM-Reference-Format}
\balance
\bibliography{sample-base}

\end{document}